%% file: PaperForReview.tex
\documentclass[10pt,twocolumn,letterpaper]{article}

\usepackage[pagenumbers]{cvpr} 

\usepackage{graphicx}
\usepackage{amsmath}
\usepackage{amssymb}
\usepackage{booktabs}
\usepackage{cite}
\usepackage{array}
\usepackage{pifont}
\usepackage[pagebackref,breaklinks,colorlinks]{hyperref}
\usepackage{multirow}
\usepackage[table]{xcolor}

\definecolor{lightpurple}{RGB}{200, 230, 245}
\definecolor{lightblue}{RGB}{230, 245, 255}

\usepackage[capitalize]{cleveref}
\crefname{section}{Sec.}{Secs.}
\Crefname{section}{Section}{Sections}
\Crefname{table}{Table}{Tables}
\crefname{table}{Tab.}{Tabs.}

\def\cvprPaperID{***} 
\def\confName{CVPR}
\def\confYear{2025}

\begin{document}

\title{Enhancing Visual In-Context Learning by Multi-Faceted Fusion}

\author{
    Wenwen Liao\textsuperscript{1} \quad
    Jianbo Yu\textsuperscript{2}\thanks{Corresponding author: {\tt\small jb\_yu@fudan.edu.cn}} \quad
    Yuansong Wang\textsuperscript{3} \quad
    Qingchao Jiang\textsuperscript{4} \quad
    Xiaofeng Yang\textsuperscript{2} \\[2mm]
    \textsuperscript{1}College of Intelligent Robotics and Advance Manufacturing, Fudan University\\
    \textsuperscript{2}School of Microelectronics, Fudan University\\
    \textsuperscript{3}Tsinghua Shenzhen International Graduate School, Tsinghua University\\
    \textsuperscript{4}School of Information Science and Engineering, East China University of Science and Technology\\
    {\tt\small wwliao24@m.fudan.edu.cn, jb\_yu@fudan.edu.cn} 
}

\twocolumn[{
\maketitle
\begin{center}
    \captionsetup{type=figure}
    \includegraphics[width=1\textwidth]{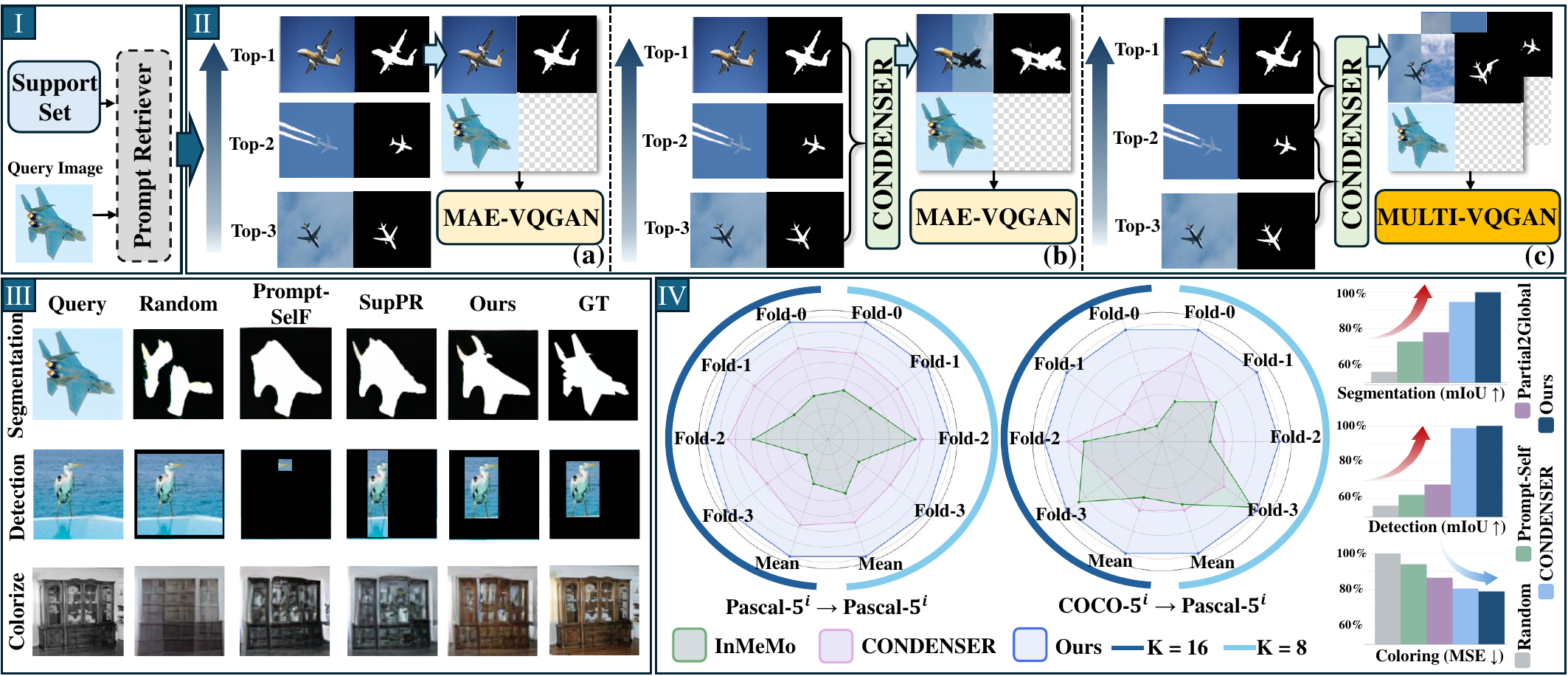}
    \captionof{figure}{\textbf{Overview and performance of our proposed framework.} \textbf{(I)} and \textbf{(II)} illustrate the core pipeline, contrasting our \textbf{collaborative fusion} with MULTI-VQGAN \textbf{(c)} against the baseline MAE-VQGAN \textbf{(a)} and the CONDENSER \textbf{(b)}. \textbf{(III)} and \textbf{(IV)} showcase the framework's superior performance through qualitative and quantitative comparisons on different tasks against existing methods.}
    \vspace{1mm}
    \label{top}
\end{center}
}]

\begin{abstract}
Visual In-Context Learning (VICL) has emerged as a powerful paradigm, enabling models to perform novel visual tasks by learning from in-context examples. The dominant “retrieve-then-prompt” approach typically relies on selecting the single best visual prompt, a practice that often discards valuable contextual information from other suitable candidates. While recent work has explored fusing the top-K prompts into a single, enhanced representation, this still simply collapses multiple rich signals into one, limiting the model's reasoning capability. We argue that a more multi-faceted, collaborative fusion is required to unlock the full potential of these diverse contexts. To address this limitation, we introduce a novel framework that moves beyond single-prompt fusion towards an multi-combination collaborative fusion. Instead of collapsing multiple prompts into one, our method generates three contextual representation branches, each formed by integrating information from different combinations of top-quality prompts. These complementary guidance signals are then fed into proposed MULTI-VQGAN architecture, which is designed to jointly interpret and utilize collaborative information from multiple sources. Extensive experiments on diverse tasks, including foreground segmentation, single-object detection, and image colorization, highlight its strong cross-task generalization, effective contextual fusion, and ability to produce more robust and accurate predictions than existing methods. Code will be released after acceptance.

\end{abstract}

\section{Introduction}
\label{sec:intro}

In-Context Learning (ICL) is first introduced in the field of Natural Language Processing (NLP) by GPT-3 \cite{brown2020language}. This paradigm enables large language models to perform new tasks by providing task demonstrations within the input, allowing them to imitate and generalize in a few-shot manner. The remarkable success of ICL \cite{touvron2023llama,touvron2023llama2,grattafiori2024llama} soon inspired researchers to extend this paradigm to the multi-modal domain \cite{peng2023kosmos,sun2024generative,wang2023context}. A representative work, Flamingo \cite{alayrac2022flamingo}, integrates interleaved images and texts as contextual examples, thereby equipping models with ICL capabilities for vision-language tasks.

Building on this foundation, the research community has further explored Visual In-Context Learning (VICL) \cite{bar2022visual}, which aims to enable models to understand and perform arbitrary visual tasks solely from given visual examples. Recent VICL models typically follow a “retrieve-then-prompt” paradigm, where a retriever module searches the support set (as shown in Figure~\ref{top} (I)) for the most similar examples and uses them as visual prompts to guide a generative model, usually MAE-VQGAN \cite{bar2022visual}, in completing the query task (as shown in Figure~\ref{top} (II a)). Within this framework, visual prompts serve as a key factor directly influencing model performance, as more relevant and higher-quality prompts generally lead to more accurate predictions, motivating a line of research on how to better select or enhance prompts. Prompt-SelF \cite{sun2025exploring} performs pixel-level retrieval strategy. VPR \cite{zhang2023makes} automates prompt selection via unsupervised nearest-neighbor search or a supervised retriever optimized for ICL performance. Partial2Global \cite{xu2024towards} refines ranking with a transformer-based list-wise ranker to identify the global optimum. Moving from selection to enhancement, InMeMo \cite{zhang2024instruct} improves instructional quality by adding lightweight, learnable perturbations to in-context pairs.

Although retrieving or ranking the most similar examples has driven progress, the single-choice paradigm remains inherently limited because selecting only one prompt inevitably discards complementary cues and restricts generalization. Methods such as CONDENSER \cite{wang2025embracing} take a natural step forward by fusing information from the top-K prompts to form a more complete context (Figure~\ref{top} (II b)). However, simply aggregating these rich signals into a single condensed representation introduces a subtler issue, since the model loses the opportunity to weigh, compare, and reconcile diverse and sometimes conflicting evidence across prompts. Therefore, naïve multi-prompt fusion is still insufficient, and unlocking the full reasoning potential of these varied contexts requires a more multi-faceted collaborative fusion mechanism.

To this end, we propose our novel framework, depicted in Figure \ref{top} (II c). Instead of producing a single condensed prompt, our approach facilitates a collaborative fusion process that integrates information from various combinations of these prompts. To create a richer and more multi-faceted contextual representation, we first introduce a new grouping strategy for top-tier prompts, termed \textbf{Multiple Prompt Group Selection (MPGS)}. Motivated by the principles of Disentangled Representation Learning (DRL)~\cite{lake2017building,bengio2013representation}, MPGS uses the similarity between in-context prompts and the query as a supervision signal to disentangle the support information into three groups: \textit{holistic}, \textit{high-similarity}, and \textit{low-similarity}. These three groups are then treated as separate representation branches and subsequently fed into our novel \textbf{MULTI-VQGAN} architecture for collaborative contextual decoding. This new generative model is specifically designed to interpret and leverage the complex, collaborative guidance from multiple sources simultaneously, allowing it to produce a more robust and accurate final prediction compared to MAE-VQGAN relying on a single input.

As illustrated in Figure~\ref{top} (III), our method produces clearer structural details and more coherent semantic regions, owing to the effective disentanglement and grouping of prompt information by MPGS, as well as the collaborative multi-branch modeling by MULTI-VQGAN. On the quantitative side (Figure~\ref{top} (IV)), our approach consistently outperforms existing methods across segmentation, detection, and colorization tasks, and maintains a significant advantage not only in the in-domain setting (Pascal-5$^i$ $\rightarrow$ Pascal-5$^i$) but also in the more challenging cross-domain scenario (COCO-5$^i$ $\rightarrow$ Pascal-5$^i$). These results collectively demonstrate the effectiveness of MPGS in contextual disentanglement and the robustness of MULTI-VQGAN in multi-source collaborative modeling, leading to strong cross-task and cross-domain generalization.

To sum up, this paper's contributions are:

\begin{itemize}

    \item \textbf{Collaborative Fusion Framework:} We introduce a multi-combination collaborative fusion paradigm that goes beyond single-prompt selection or naive single-group condensing, forming richer and more structured contextual representations.

    \item \textbf{A Novel Grouping Strategy:} We propose MPGS to disentangle top-tier prompts into holistic, high-similarity, and low-similarity groups based on prompt--query similarity, enabling complementary contextual cues to be explicitly preserved and utilized.

    \item \textbf{Multi-Branch Fusing Architecture:} We design MULTI-VQGAN to jointly interpret and fuse the three representation branches, allowing effective multi-source contextual integration and more robust generation.

    \item \textbf{Extensive Empirical Validations:} Experiments across segmentation, detection, and colorization tasks demonstrate clear performance gains and strong cross-task and cross-domain generalization.
\end{itemize}


\begin{figure*}
  \centering
  \vspace{-3mm} 
  \includegraphics[width=\linewidth]{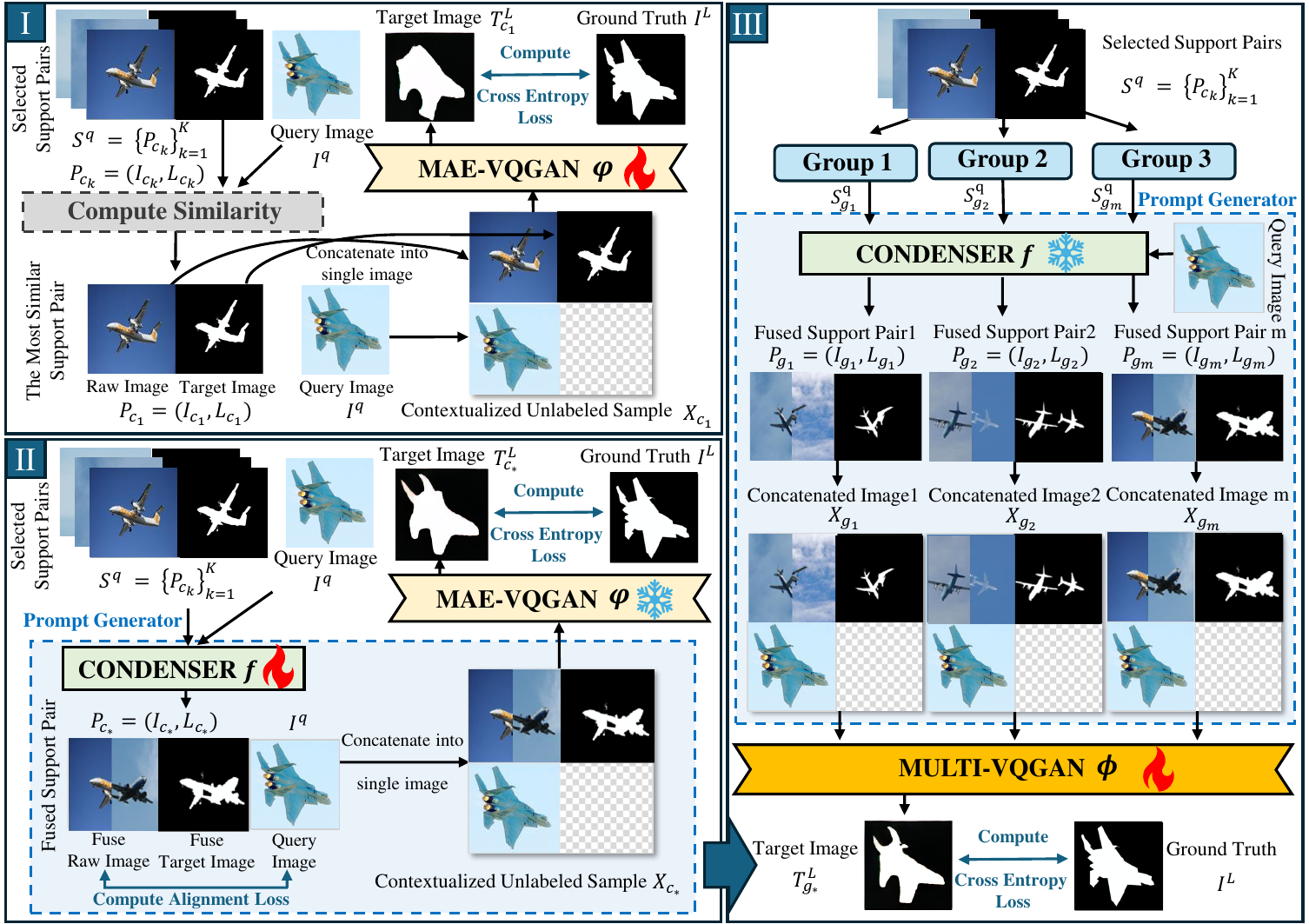} 

  \vspace{-1mm} 
  \caption{\textbf{Pipeline of our framework in contrast to existing methods.} 
\textbf{(I)} The single prompt method uses only the most similar prompt to guide a MAE-VQGAN. 
\textbf{(II)} The CONDENSER method trains a module to fuse multiple prompts into one, which then guides a frozen MAE-VQGAN. 
\textbf{(III)} Our proposed MULTI-VQGAN architecture. The support pairs are grouped by MPGS, then processed by the frozen prompt generator, and the resulting multiple fused prompts are used to train the MULTI-VQGAN for a more robust, hierarchical fusion.}
  \label{fig:ppl}
  \vspace{-5mm} 
\end{figure*}

\section{Preliminary}
\label{sec:mae-vqgan}

The procedure in existing VICL works \cite{sun2025exploring, xu2024towards, zhang2024instruct, zhang2023makes} is shown in Figure \ref{fig:ppl} (I), let the query set be \(Q = \{I^q, L^q\}_{q=1}^{N_q}\), where \(I^q\) and \(L^q\) denote the query image and its corresponding ground-truth label (e.g., an image annotated with bounding boxes or segmentation masks), and \(I^q, L^q \in \mathbb{R}^{H_0/2 \times W_0/2 \times 3}\), with \(H_0 = W_0 = 224\). Given a query image \(I^q\) sampled from \(Q\), a subset of relevant prompts is selected from the database based on image-image similarity, denoted as \(S^q = \{P_{c_k}\}_{k=1}^{K}\), where each prompt \(P_i = (I_i, L_i)\) is an image-label pair. Consistent with the competitive prompt selection strategy adopted by most existing methods, only the top-ranked prompt \(P_{c_1}\) is used to construct the contextualized sample, defined as  
\begin{equation}
X_{c_1} = 
\begin{bmatrix}
I_{c_1} & L_{c_1} \\
I^q & \; L[\text{MASK}] \;
\end{bmatrix}\in \mathbb{R}^{H_0 \times W_0 \times 3},
\end{equation}

\noindent where \(L[\text{MASK}]\) denotes the padding mask associated with the query image. The constructed \(X_{c_1}\) is then fed into MAE-VQGAN $\varphi$ \cite{bar2022visual} to reconstruct the \(L[\text{MASK}]\) region.

\begin{equation}
  T_{c_1} = \varphi(X_{c_1})=\begin{bmatrix}
I_{c_1} & L_{c_1} \\
I^q & \; T^L_{c_1} \;
\end{bmatrix}. 
\label{eq:taeget}
\end{equation}

The cross-entropy loss compares the target image \(T_{c_1}^L\) with the ground truth label \(I^L\) of the query image, expressed as
\begin{equation}
\mathcal{L}_{\text{CE}} = - \sum_{j=1}^d I^L_j \log T^L_{c_1,j},
\label{eq:ce}
\end{equation}
where $d$ is the feature dimension.

\section{Method}
\label{sec:me}

\subsection{Overview}
The overall pipeline of our proposed framework is illustrated in Figure~\ref{fig:ppl} (II and III). First, a Prompt Generator is designed to fuse multiple prompts into a coherent representation (Section~\ref{sec:prompt generator}). 
Second, we group the candidate support prompts according to MPGS. Each group is independently processed by the frozen Prompt Generator. The resulting outputs are then collectively fed into our novel MULTI-VQGAN architecture, an enhanced model based on MAE-VQGAN, which is then trained to integrate these multiple fused representations for the final prediction (Section~\ref{sec:multi}). Each stage is detailed in the following sections.

\subsection{Training of Prompt Generator}
\label{sec:prompt generator}

Rather than using only the top-1 prompt, we fuse multiple similar prompts into a unified one, as shown in Figure~\ref{fig:ppl} (II), by first training the \textbf{Prompt Generator} $\Phi_{pg}$. It takes as input a set of selected support pairs $S^q$ together with a query image \(I^q\), and outputs a contextualized unlabeled sample \(X_{c_*}\). 

\begin{equation}
X_{c_*} = \Phi_{pg}(S^q,I^q).
\end{equation}

Specifically, the core component of the Prompt Generator is the \textbf{CONDENSER} \(f\) \cite{wang2025embracing}, which fuses multiple support pairs into a single fused support pair, denoted as $P_{c_*} = (I_{c_*}, L_{c_*})$, where \(I_{c_*}\) is the fused image and \(L_{c_*}\) the corresponding fused label.

\begin{equation}
P_{c_*} = f(S^q, I^q).   
\end{equation}

Instead of the top-ranked prompt \(P_{c_1}\), the Fused Support Pair $P_{c_*}$ is used to construct the contextualized sample, defined as  

\begin{equation}
X_{c_*} = 
\begin{bmatrix}
{I}_{c_*} & L_{c_*} \\
I^q & \; L[\text{MASK}] \;
\end{bmatrix}.
\end{equation}

During the training phase, the parameters of the MAE-VQGAN model \(\varphi\) are kept frozen. The contextualized sample \(X_{c_*}\) is fed into the frozen MAE-VQGAN \(\varphi\), which produces an target image \(T_{c_*}^L\) as defined in Equation~(\ref{eq:taeget}).

The training of the \(f\) is driven by two objectives: an alignment loss and a cross-entropy loss. The alignment loss, which ensures that the fused prompt remains semantically consistent with the query image, is defined as:
\begin{equation}
\mathcal{L}_{\text{align}} = \| I_{c_*} - I^q \|_2^2.
\end{equation}

Meanwhile, the cross-entropy loss $\mathcal{L}_{\text{CE}}$ compares \(T_{c_*}^L\) with the ground truth label \(I^L\) of the query image, calculated as expressed in Equation~(\ref{eq:ce}).

The overall training objective is then given by:
\begin{equation}
\mathcal{L} =  (1-\lambda )\mathcal{L}_{\text{align}} + \lambda \, \mathcal{L}_{\text{CE}},
\label{loss_pg}
\end{equation}
where \(\lambda\) is a hyperparameter that balances the alignment and cross-entropy terms. The gradient derived from this objective is backpropagated exclusively through the CONDENSER \(f\), thereby updating its parameters and progressively enhancing its ability to generate effective fused prompts.

\subsection{Training of MULTI-VQGAN}
\label{sec:multi}

\subsubsection{Multiple Prompt Group Selection}

According to the principles of DRL \cite{lake2017building,bengio2013representation}, different support pairs in $S^q$ contribute distinct types of information: highly similar pairs provide critical fine-grained cues, while less similar ones offer valuable variance, contextual diversity, and robustness. Moreover, \cite{locatello2019challenging,locatello2019disentangling} have theoretically shown that pure unsupervised DRL is impossible without introducing inductive biases into the model or the data. In other words, without structural constraints, a model cannot automatically learn to disentangle latent variation factors. Therefore, directly merging all support pairs in $S^q$ as CONDENSER \cite{wang2025embracing} risks noisy features overshadowing the discriminative details.

To address this issue, we propose \textbf{Multiple Prompt Group Selection (MPGS)}, a multi-branch fusion strategy that partitions the support pairs into groups for representation disentanglement. Formally, given $K$ selected support pairs $S^q$ we use similarity as the DRL supervision signal to partition them into three groups. The holistic group $S^q_{g_m}$ contains all support pairs and provides complete contextual information. The high-similarity group $S^q_{g_1}$ consists of the top $K_{g_1}$ most relevant support pairs and serves as the \textit{core variation factor}, capturing salient fine-grained cues that are directly correlated with the query. Conversely, the low-similarity group $S^q_{g_2}$ is formed from the last $K_{g_2}$ pairs and captures \textit{interfering variation factors}. This group emphasizes more general and contrastive features, enhancing robustness and reducing overfitting to top matches, a strategy shown to encourage diverse representation learning~\cite{lee2018diverse}. By disentangling support information, MPGS improves the explainability, generalizability, and controllability of the final representation, consistent with findings in representation disentanglement and clustering-based learning~\cite{wu2020improving,chen2016infogan}.

\begin{figure*}
  \centering
  \vspace{-3mm} 
  \includegraphics[width=\linewidth]{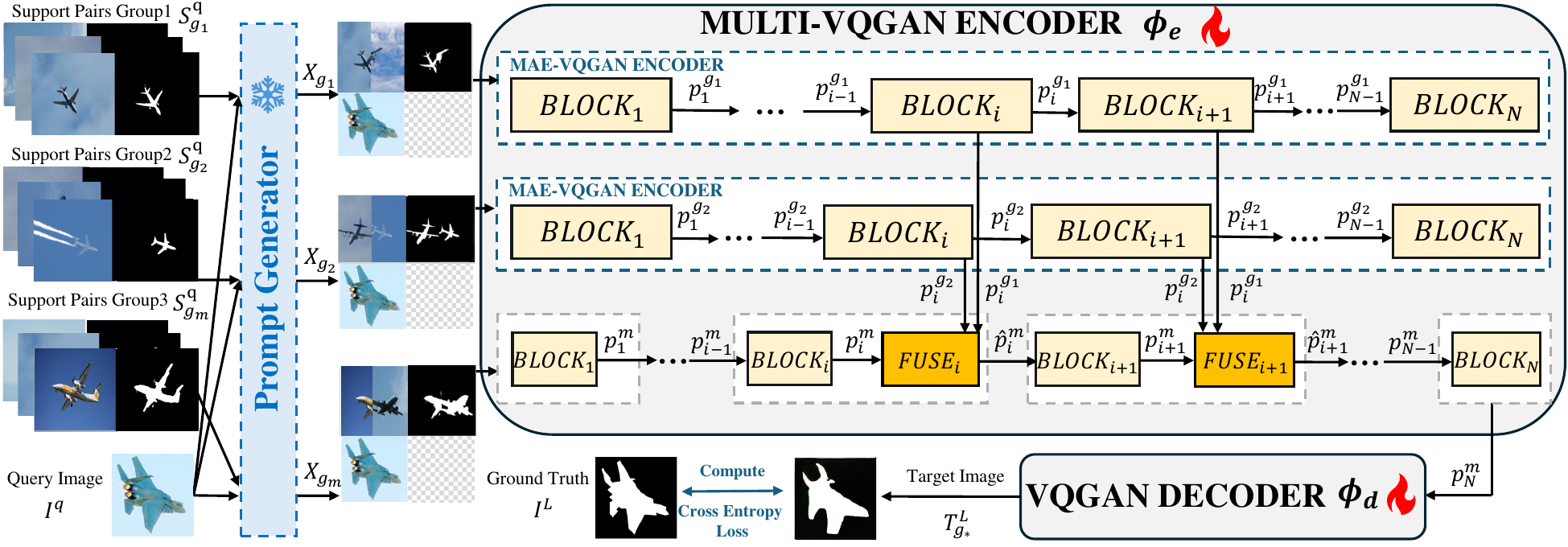} 

  \vspace{-1mm} 
  \caption{The architecture of MULTI-VQGAN, which performs hierarchical feature fusion with a multi-branch encoder.}
  \label{fig:multi}
  \vspace{-5mm} 
\end{figure*}

\subsubsection{MULTI-VQGAN for Hierarchical Fusion}

A key challenge after disentangling the support set is how to structurally integrate these separated feature groups. Drawing on the principles of Hierarchical Disentangled Representation Learning (Hierarchical DRL) \cite{li2020progressive,ross2021benchmarks,tong2019hierarchical}, which suggest that generative factors reside at different semantic abstraction levels, we regard the block layers in MAE-VQGAN as corresponding to distinct abstraction stages within its generative process.

To effectively integrate this multi-level knowledge from three groups, we propose a customized fusion-generation architecture, termed MULTI-VQGAN, as illustrated in Figure~\ref{fig:multi}. This model extends the standard MAE-VQGAN into a multi-branch encoder–decoder framework designed specifically for hierarchical, block-wise feature fusion. It aims to align the semantic hierarchy of our disentangled features with the architectural hierarchy of the encoder.
 
As illustrated in Figure~\ref{fig:ppl} (III), given three prompt groups processed by MPGS, they are first passed through the frozen Prompt Generator \(\Phi_{pg}\) to produce concatenated images:
\begin{align}
X_{g_1} &= \Phi_{pg}(S^q_{g_1}, I^q), \\
X_{g_2} &= \Phi_{pg}(S^q_{g_2}, I^q), \\
X_{g_m} &= \Phi_{pg}(S^q_{g_m}, I^q).
\end{align}

The concatenated images from the three groups are then fed into our proposed MULTI-VQGAN architecture, which is designed to perform a hierarchical, block-wise fusion of these parallel information streams. As illustrated in Figure~\ref{fig:multi}, the core of this model is a multi-branch encoder \(\phi_e\), which processes each group's input to progressively integrate their features.

The process begins with the two guiding inputs, \(X_{g_1}\) and \(X_{g_2}\), which are passed into two auxiliary sub-branches. These branches are identical, pre-trained MAE-VQGAN encoders, to extract a hierarchy of guidance features. The output feature vector from the \(i\)-th block of these branches is denoted as:
\begin{equation}
p_i^{g_1} = {BLOCK}_i(p_{i-1}^{g_1}) \quad \text{and} \quad p_i^{g_2} = {BLOCK}_i(p_{i-1}^{g_2}),
\end{equation}
where \(p_0^{g_1}\) and \(p_0^{g_2}\) represent initial patch embedding of \(X_{g_1}\) and \(X_{g_2}\).

Concurrently, the holistic input \(X_{g_m}\) is processed by the mainstream branch, which serves as our generative backbone composed of MAE-VQGAN encoder blocks and FUSE modules. 
Motivated by hierarchical representation learning \cite{johnson2016perceptual}, we note that intermediate layers form a critical stage where high-level semantics remain adaptable while low-level details are already established. Thus, we hypothesize that fusing at mid-level layers offers an effective balance between structural guidance and fine-grained refinement, enabling the model to leverage both global contextual cues and localized appearance information more effectively.

Accordingly, fusion is applied across a continuous intermediate range of blocks, from \(N_{\text{down}}\) to \(N_{\text{up}}\), within the backbone. 
Specifically, following each block within this designated range, a learnable FUSE module integrates the mainstream features with corresponding guidance features from the auxiliary branches. 
Formally, this fusion process is defined as:
\begin{equation}
\hat{p}_i^m = \text{FUSE}_i\left(\text{BLOCK}_i(\hat{p}_{i-1}^m),\, p_i^{g_1},\, p_i^{g_2}\right), 
i \in [N_{\text{down}},\, N_{\text{up}}],
\end{equation}
where \(\hat{p}_i^m\) represents the enriched feature vector post-fusion at level \(i\), and \(\hat{p}_0^m\) is the initial patch embedding of \(X_{g_m}\). 
This updated vector, \(\hat{p}_i^m\), is then passed as input to the subsequent block in the sequence.

By progressively injecting guidance from two sub-branches, the mainstream representation is incrementally refined into a more contextually enriched form. After all \(N\) blocks, the final representation is obtained as:
\begin{equation}
{p}_N^m = {BLOCK}_N({p}_{N-1}^m).
\end{equation}

The fused feature vector is then passed into the activated VQGAN decoder \(\phi_d\) to produce the final output image:
\begin{equation}
T_{g*} = \phi_d({p}_N^m)=\begin{bmatrix}
I_{c_1} & L_{c_1} \\
I^q & \; T^L_{g_*} \;
\end{bmatrix}.
\end{equation}

The model is trained end-to-end by computing the cross-entropy loss $\mathcal{L}_{\text{CE}}$ between the generated taeget image \(T_{g*}^L\) and the ground-truth image \(I^L\), calculated as expressed in Equation~(\ref{eq:ce}). By injecting each feature from auxiliary sub-branches to holistic branch at its corresponding architectural depth, this hierarchical design prevents the re-entanglement of information and facilitates a more controllable and robust synthesis of the final representation.

\begin{figure}
  \centering
  \includegraphics[width=\linewidth]{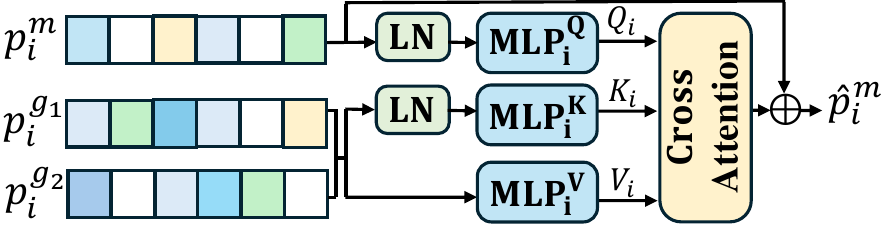} 
  \caption{The architecture of the FUSE module.}
  \label{fig:fuse}
  \vspace{-1em}  
\end{figure}

\subsubsection{FUSE Module Architecture}

The FUSE module, illustrated in Figure \ref{fig:fuse}, is built upon a \textbf{Cross-Attention Mechanism} that enables dynamic, context-aware feature integration from multiple sources.  

The core idea lies in the formulation of the Query (Q), Key (K), and Value (V): the mainstream feature serves as the Query, while the high- and low-similarity guidance features jointly provide the Key and Value. 

After the mainstream feature \(\hat{p}_{i-1}^m\) passes through the frozen backbone block \({BLOCK}_i\) to produce \(p_i^m\), the fusion is performed as:
\begin{align}
Q_i &= \text{MLP}_i^Q\!\left(\text{LN}(p_i^m)\right), \\
K_i &= \text{MLP}_i^K\!\left(\text{LN}\!\left([p_i^{g_1}; p_i^{g_2}]\right)\right), \\
V_i &= \text{MLP}_i^V\!\left([p_i^{g_1}; p_i^{g_2}]\right),
\end{align}
where $[\,\cdot\,;\,\cdot\,]$ denotes feature concatenation. Layer Normalization (LN) stabilizes feature distributions, and three learnable MLPs ($\text{MLP}_i^Q$, $\text{MLP}_i^K$, $\text{MLP}_i^V$) project them into the attention space.

A standard cross-attention operation then computes attention weights between the mainstream Query and guidance Key, producing a weighted sum of Value vectors. The result is integrated into the mainstream feature via a residual connection:
\begin{equation}
\hat{p}_i^m = p_i^m + \text{CrossAttention}(Q_i, K_i, V_i)    
\end{equation}

This mechanism enables the mainstream branch to dynamically query the guidance sub-branches and selectively integrate fine-grained and contextual information, rather than performing static feature fusion. 

\begin{table*}
\centering
\small
\begin{tabular}{p{1cm}lccccccc}
\toprule
\textbf{Type} & \textbf{Model} & \multicolumn{5}{c}{\textbf{Seg. (mIoU $\uparrow$)}} & \textbf{Det. (mIoU $\uparrow$)} & \textbf{Col. (MSE $\downarrow$)} \\
\cmidrule(lr){3-7}
& & \textbf{Fold-0} & \textbf{Fold-1} & \textbf{Fold-2} & \textbf{Fold-3} & \textbf{Mean} & & \\
\midrule\midrule
\multirow{5}{*}{\shortstack[l]{Single\\Prompt\\Selection}} 
& Random \cite{bar2022visual} & 28.66 & 30.21 & 27.81 & 23.55 & 27.56 & 25.45 & 0.67 \\
& UnsupPR \cite{zhang2023makes} & 34.75 & 35.92 & 32.41 & 31.16 & 33.56 & 26.84 & 0.63 \\
& SupPR \cite{zhang2023makes} & 37.08 & 38.43 & 34.40 & 32.32 & 35.56 & 28.22 & 0.63 \\
& Prompt-Self \cite{sun2025exploring} & 35.69 & 38.25 & 35.86 & 33.37 & 35.79 & 28.08 & 0.63 \\
& Partial2Global \cite{xu2024towards} & 38.81 & 41.54 & 37.25 & 36.01 & 38.40 & 30.66 & 0.58 \\
& InMeMo \cite{zhang2024instruct} & 41.65 & 47.68 & 42.43 & 40.80 & 43.14 & 43.21 & - \\
& Task-Level Prompting \cite{zhu2025exploring} & 39.09{\tiny\,±0.77} & 44.37{\tiny\,±0.98} & 37.93{\tiny\,±0.42} & 32.40{\tiny\,±1.06} & 38.45 & 29.03{\tiny\,±2.84} & 0.62{\tiny\,±1.27} \\
\midrule
\multirow{2}{*}{Voting} 
& Prompt-Self$_{\text{w/ voting}}$ \cite{sun2025exploring} & 42.48 & 43.34 & 39.76 & 38.50 & 41.02 & 29.83 & - \\
& Partial2Global$_{\text{w/ voting}}$ \cite{xu2024towards} & 43.23 & 45.50 & 41.79 & 40.22 & 42.69 & 32.52 & - \\
\midrule
\multirow{4}{*}{\cellcolor{white}Fusion} 
& CONDENSER$_{K=8}$ \cite{wang2025embracing} 
& 44.63 & 50.59 & 42.98 & 43.82 & 45.51 & 41.36 & 0.56 \\
& \cellcolor{lightblue}Ours$_{K=8}$ 
& \cellcolor{lightblue}\underline{47.13}{\tiny\,±0.06} 
& \cellcolor{lightblue}\underline{53.83}{\tiny\,±0.09} 
& \cellcolor{lightblue}\underline{45.05}{\tiny\,±0.07} 
& \cellcolor{lightblue}\underline{47.34}{\tiny\,±0.10} 
& \cellcolor{lightblue}\underline{48.34}
& \cellcolor{lightblue}43.67{\tiny\,±0.07} 
& \cellcolor{lightblue}\underline{0.54}{\tiny\,±0.28} \\
& CONDENSER$_{K=16}$ \cite{wang2025embracing} & 45.53 & 52.06 & 44.33 & 44.58 & 46.63 & \underline{44.64} & \underline{0.54} \\
& \cellcolor{lightpurple}Ours$_{K=16}$ 
& \cellcolor{lightpurple}\textbf{47.64}{\tiny\,±0.06} 
& \cellcolor{lightpurple}\textbf{54.96}{\tiny\,±0.09} 
& \cellcolor{lightpurple}\textbf{46.07}{\tiny\,±0.07} 
& \cellcolor{lightpurple}\textbf{48.38}{\tiny\,±0.10} 
& \cellcolor{lightpurple}\textbf{49.26}
& \cellcolor{lightpurple}\textbf{45.19}{\tiny\,±0.08} 
& \cellcolor{lightpurple}\textbf{0.53}{\tiny\,±0.25} \\
\bottomrule
\end{tabular}
\caption{\textbf{Performance comparison across different tasks.} \textbf{Bold} indicates the best performance, and \underline{underlined} indicates the second-best.}
\label{tab:main}
\end{table*}

\begin{table}
\centering
\footnotesize
\setlength{\tabcolsep}{3pt}
\begin{tabular}{p{3cm}p{0.78cm}p{0.78cm}p{0.78cm}p{0.78cm}p{0.78cm}}
\toprule
\textbf{Model} & \multicolumn{5}{c}{\textbf{Seg. (mIoU $\uparrow$)}} \\
\cmidrule(lr){2-6}
& \textbf{Fold-0} & \textbf{Fold-1} & \textbf{Fold-2} & \textbf{Fold-3} & \textbf{Mean} \\
\midrule\midrule
InMeMo \cite{zhang2024instruct} & 38.74 & 43.82 & 40.45 & 37.12 & 40.03 \\
Prompt-SelF \cite{sun2025exploring} & 40.13 & 42.14 & 37.84 & \textbf{38.52} & 39.66 \\
CONDENSER$_{K=8}$ \cite{wang2025embracing} & 40.50 & 43.59 & 40.95 & 36.01 & 40.26 \\
\rowcolor{lightblue}
Ours$_{K=8}$ & \underline{41.40} & \underline{45.78} & \textbf{42.98} & \underline{37.31} & \underline{41.87} \\
CONDENSER$_{K=16}$ \cite{wang2025embracing} &  40.37 & 44.85 & 41.03 & 35.84 & 40.52 \\
\rowcolor{lightpurple}
Ours$_{K=16}$ & \textbf{42.38} & \textbf{47.73} & \underline{41.86} & 37.60 & \textbf{42.39} \\
\bottomrule
\end{tabular}
\caption{\textbf{Cross-dataset performance evaluation.} We train models on COCO-5$^i$ and test on Pascal-5$^i$ for the segmentation task.}
\label{tab:coco}
\end{table}

\begin{table}
\centering
\footnotesize
\setlength{\tabcolsep}{5pt}
\begin{tabular}{p{0.7cm}cccccc} 
\toprule
\textbf{Groups} & \multicolumn{5}{c}{\textbf{Seg. (mIoU $\uparrow$)}} & \textbf{Det.} \\
\cmidrule(lr){2-6}
& \textbf{Fold-0} & \textbf{Fold-1} & \textbf{Fold-2} & \textbf{Fold-3} & \textbf{Mean} & \textbf{(mIoU $\uparrow$)}\\ 
\midrule\midrule
1 + 1 & \underline{47.59} & 53.60 & 44.94 & 47.42 & 48.39 & 42.29 \\
\rowcolor{lightpurple}
1 + 2 & \textbf{47.64} & \textbf{54.96} & \textbf{46.07} & \textbf{48.38} & \textbf{49.26} & \textbf{45.19} \\
1 + 4 & 46.89 & \underline{54.60} & 45.47 & \underline{47.63} & \underline{48.65} & 41.53 \\
1 + 8 & 46.54 & 53.71 & \underline{45.59} & 46.90 & 48.19 & \underline{44.86} \\
1 + 16 & 45.98 & 53.34 & 45.53 & 46.21 & 47.77 & 43.02 \\
\bottomrule
\end{tabular}
\caption{\textbf{Ablation on group number in MPGS.} The dual-branch setting (1 main + 2 auxiliary) performs best.}
\label{tab:group}
\end{table}

\begin{table*}
\centering
\small
\setlength{\tabcolsep}{11.5pt}
\begin{tabular}{clccccccc}
\toprule
\textbf{ID} & \textbf{Model} & \multicolumn{5}{c}{\textbf{Seg. (mIoU $\uparrow$)}} & \textbf{Det. (mIoU $\uparrow$)} \\
\cmidrule(lr){3-7}
& & \textbf{Fold-0} & \textbf{Fold-1} & \textbf{Fold-2} & \textbf{Fold-3} & \textbf{Mean} & \\
\midrule\midrule
\rowcolor{lightpurple}
(1) & Ours$_{K=16}$ & 
\textbf{47.64}{\tiny\,±0.06} & 
\textbf{54.96}{\tiny\,±0.09} & 
\textbf{46.07}{\tiny\,±0.07} & 
\textbf{48.38}{\tiny\,±0.10} & 
\textbf{49.26} & 
\textbf{45.19}{\tiny\,±0.08} \\
\midrule\midrule
(2) & Only $g_1$ & 
47.34{\tiny\,±0.06} & 
53.43{\tiny\,±0.10} & 
45.51{\tiny\,±0.08} & 
47.70{\tiny\,±0.10} & 
48.50 & 
44.13{\tiny\,±0.08} \\
(3) & Only $g_2$ & 
46.89{\tiny\,±0.07} & 
53.20{\tiny\,±0.10} & 
45.11{\tiny\,±0.09} & 
44.80{\tiny\,±0.12} & 
47.50 & 
42.60{\tiny\,±0.08} \\
(4) & $g_1$ as main & 
47.10{\tiny\,±0.06} & 
54.43{\tiny\,±0.10} & 
45.56{\tiny\,±0.08} & 
\underline{48.28}{\tiny\,±0.10} & 
48.84 & 
43.71{\tiny\,±0.07} \\
(5) & $g_2$ as main & 
47.20{\tiny\,±0.07} & 
54.08{\tiny\,±0.10} & 
45.42{\tiny\,±0.08} & 
47.37{\tiny\,±0.10} & 
48.52 & 
43.19{\tiny\,±0.08} \\
(6) & Random Guidance & 
\underline{47.60}{\tiny\,±0.06} & 
53.46{\tiny\,±0.10} & 
44.50{\tiny\,±0.08} & 
47.17{\tiny\,±0.11} & 
48.18 & 
42.80{\tiny\,±0.09} \\
(7) & Freeze MAE-VQGAN & 
44.98{\tiny\,±0.08} & 
51.16{\tiny\,±0.12} & 
42.88{\tiny\,±0.09} & 
44.30{\tiny\,±0.12} & 
45.83 & 
40.11{\tiny\,±0.10} \\
\midrule\midrule
(8) & 1 MLP & 
46.09{\tiny\,±0.08} & 
54.11{\tiny\,±0.10} & 
\underline{45.68}{\tiny\,±0.08} & 
47.51{\tiny\,±0.10} & 
48.35 & 
\underline{44.74}{\tiny\,±0.08} \\
(9) & 2 MLP & 
\underline{47.60}{\tiny\,±0.06} & 
54.09{\tiny\,±0.10} & 
44.56{\tiny\,±0.09} & 
46.44{\tiny\,±0.10} & 
48.17 & 
43.96{\tiny\,±0.08} \\
(10) & w/o Cross Attention & 
47.58{\tiny\,±0.06} & 
54.37{\tiny\,±0.10} & 
45.60{\tiny\,±0.08} & 
48.20{\tiny\,±0.10} & 
\underline{48.94} & 
44.73{\tiny\,±0.08} \\
(11) & w/o Residual & 
47.41{\tiny\,±0.06} & 
\underline{54.84}{\tiny\,±0.09} & 
45.60{\tiny\,±0.08} & 
47.47{\tiny\,±0.10} & 
48.83 & 
42.24{\tiny\,±0.09} \\
\bottomrule
\end{tabular}
\caption{\textbf{Structural ablation of MULTI-VQGAN.} The best scores are marked in \textbf{bold} and the second-best scores are \underline{underlined}.}
\label{tab:ABLATION}
\end{table*}

\begin{figure*}
\centering
\begin{subfigure}[t]{0.245\textwidth}
    \includegraphics[width=\linewidth]{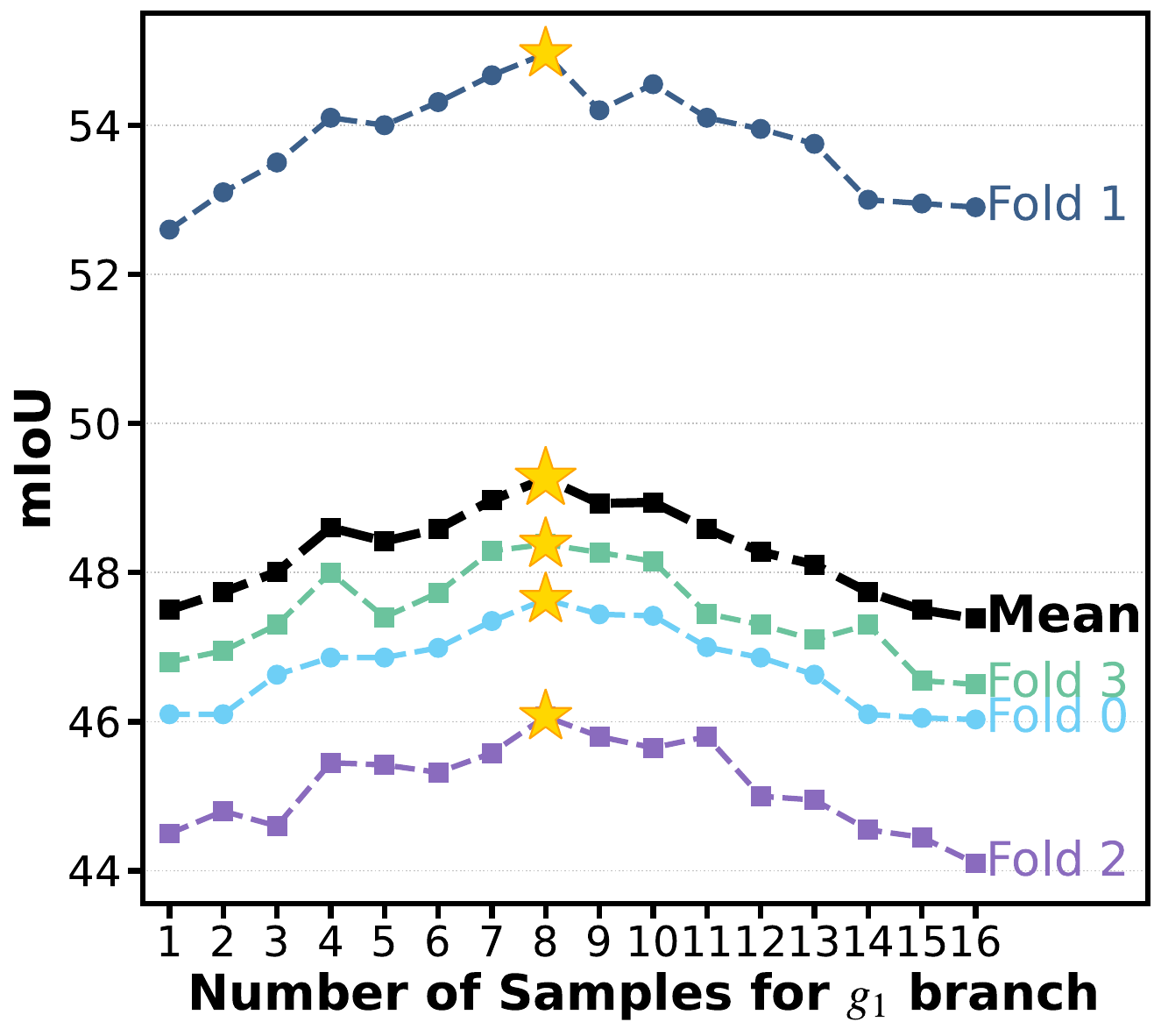}
    \caption{}
    \label{subfig:g_1}
\end{subfigure}
\hfill
\begin{subfigure}[t]{0.245\textwidth}
    \includegraphics[width=\linewidth]{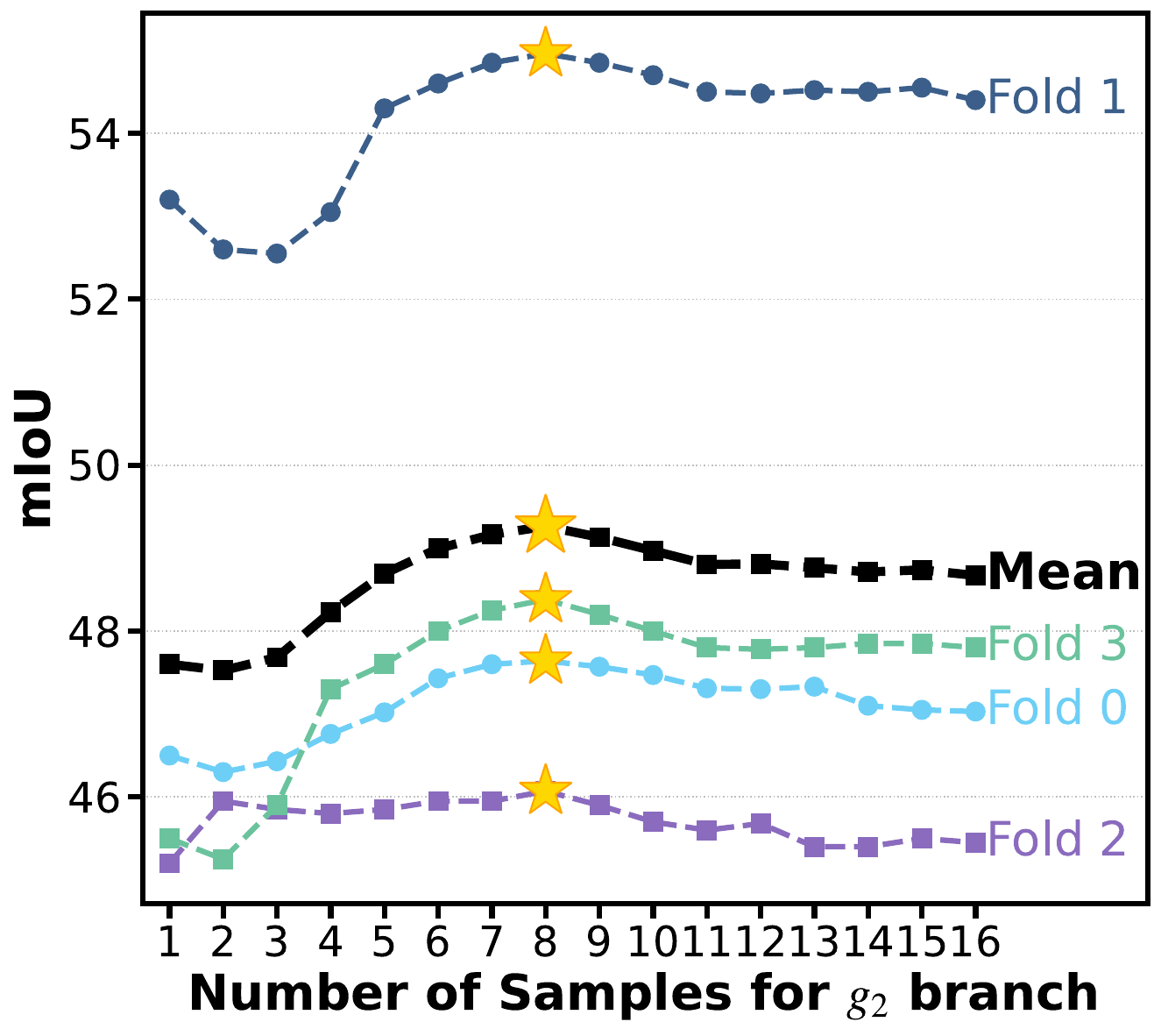}
    \caption{}
    \label{subfig:g_2}
\end{subfigure}
\hfill
\begin{subfigure}[t]{0.245\textwidth}
    \includegraphics[width=\linewidth]{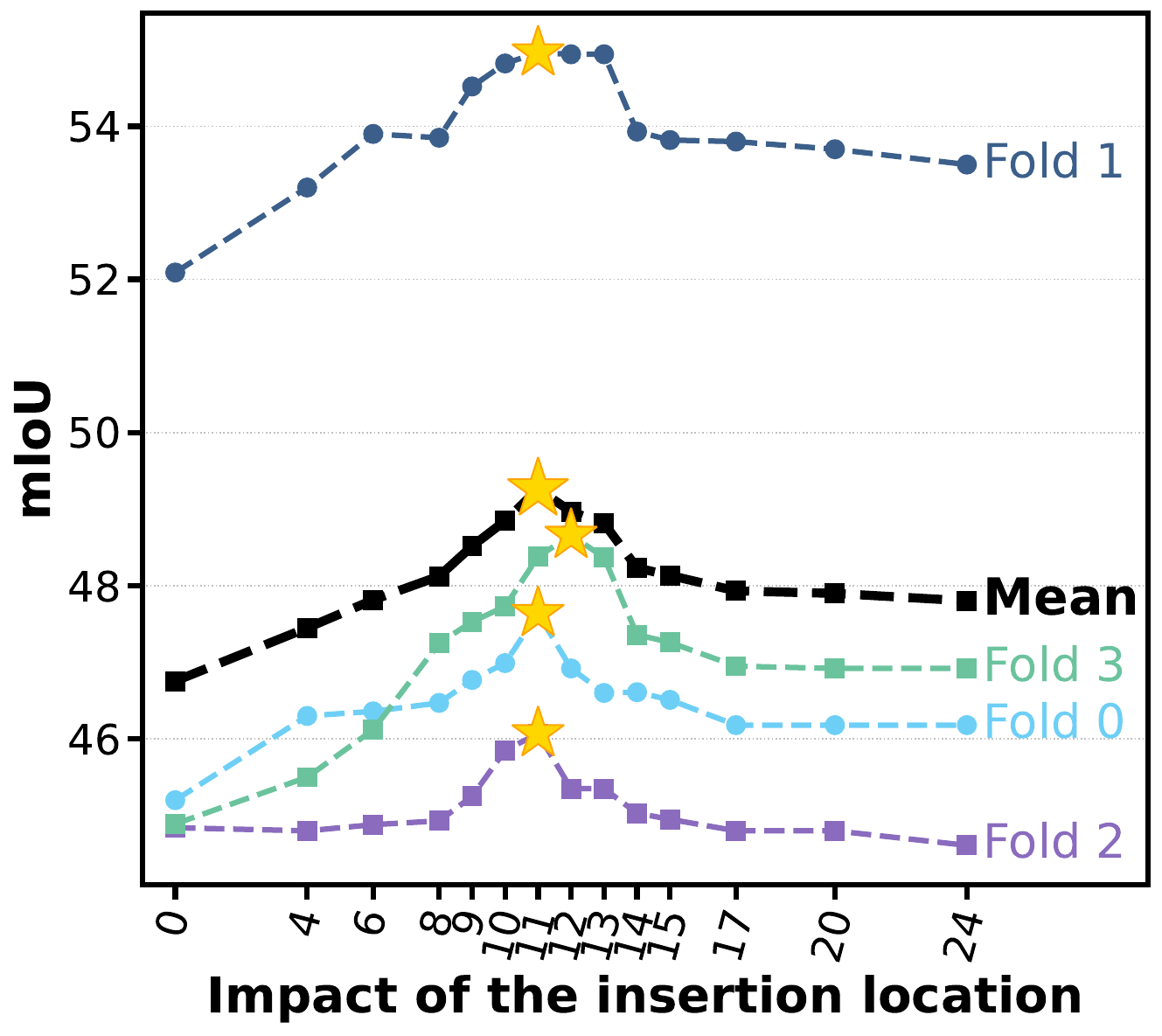}
    \caption{}
    \label{subfig:g_3}
\end{subfigure}
\hfill
\begin{subfigure}[t]{0.245\textwidth}
    \includegraphics[width=\linewidth]{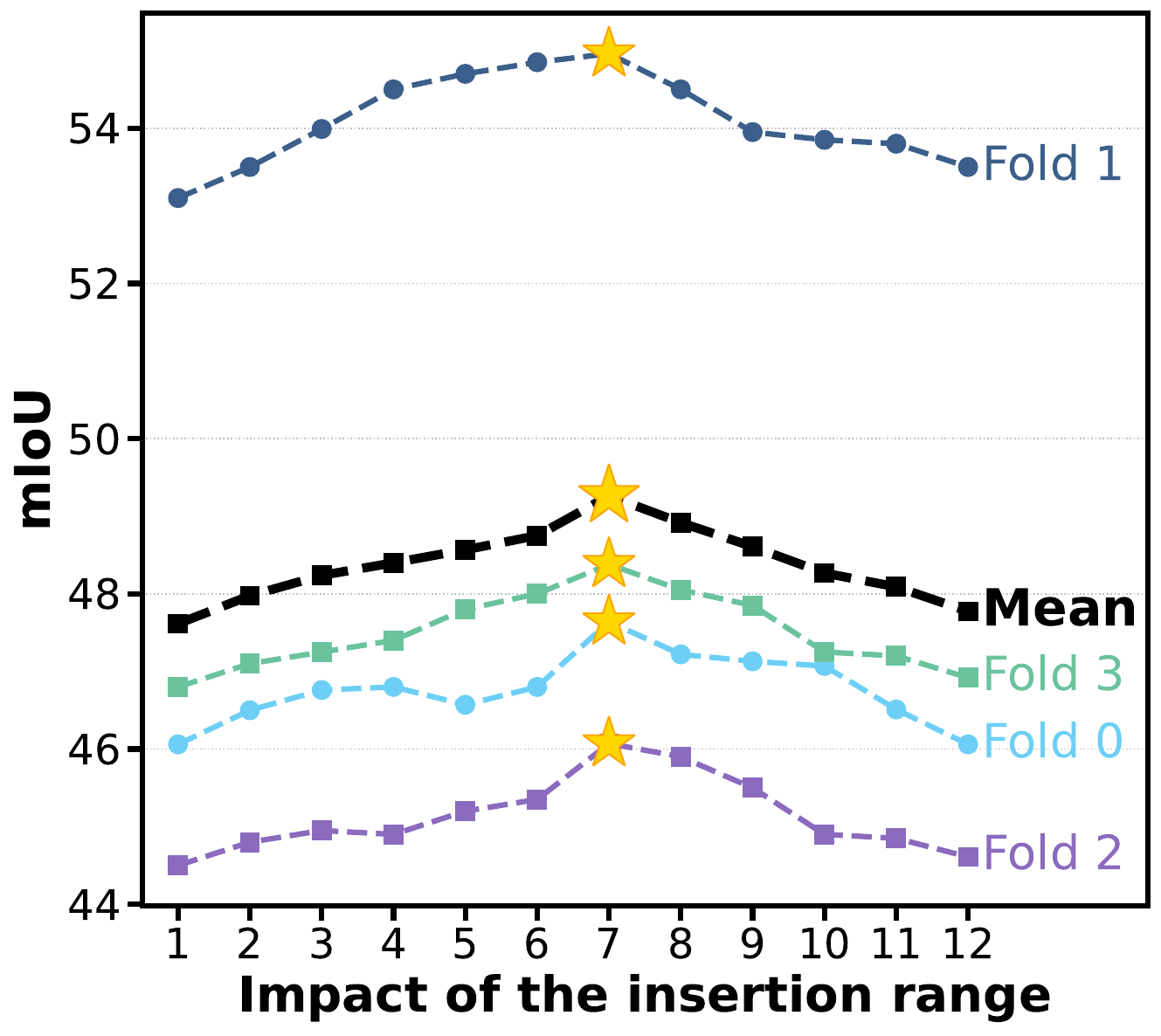}
    \caption{}
    \label{subfig:g_4}
\end{subfigure}
\caption{Ablation study on the key hyperparameters of the MULTI-VQGAN framework.}
\label{fig:ablations}
\end{figure*}

\begin{figure*}
  \centering
  \vspace{-3mm} 
  \includegraphics[width=\linewidth]{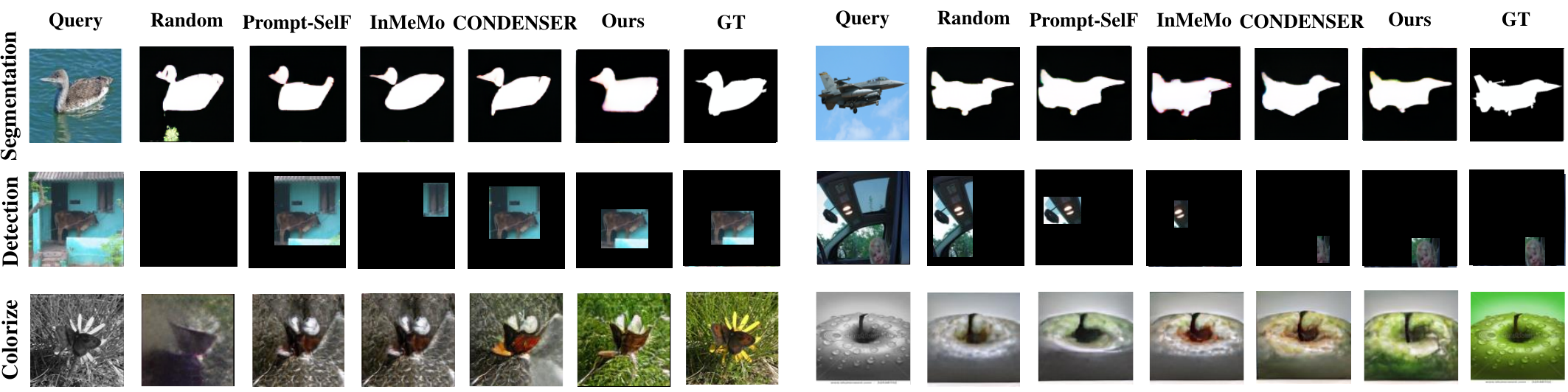} 

  \vspace{-1mm} 
  \caption{Qualitative comparison with SOTA on segmentation, detection, and colorization.}
  \label{fig:visual}
  \vspace{-5mm} 
\end{figure*}

\section{Experiments}
\label{sec:formatting}
\subsection{Datasets and Implementation Details}

We evaluate our framework on three public datasets, PASCAL-5$^i$ \cite{shaban2017one}, PASCAL VOC 2012 \cite{everingham2015pascal}, and ImageNet-1K \cite{russakovsky2015imagenet}, corresponding to foreground segmentation, single object detection, and image colorization, respectively. Further details regarding the datasets and implementation are deferred to the supplementary material.

\subsection{Baselines}

We conduct a comprehensive comparison between our method and representative baselines, which can be grouped into three categories: 
(i) \textbf{Single Prompt Selection}: methods that rely on selecting or learning a single representative prompt, including random selection~\cite{bar2022visual}, VPR~\cite{zhang2023makes} with optimized prompt retrievers, Prompt-SelF \cite{sun2025exploring} with direct prompt ranking, Partial2Global~\cite{xu2024towards} with a hierarchical prompt ranker, InMeMo~\cite{zhang2024instruct} employing learnable visual prompts, and Task-Level Prompting~\cite{zhu2025exploring} that discovers task-specific prompts; 
(ii) \textbf{Voting}: ensemble-based approaches such as Prompt-SelF~\cite{sun2025exploring} and Partial2Global~\cite{xu2024towards}, which combine outputs from different spatial arrangements of query–prompt pairs on the input canvas; 
(iii) \textbf{Prompt Fusion}: \textsc{CONDENSER} \cite{wang2025embracing}, a lightweight external plugin that compresses contextual cues from multiple prompts.

\subsection{Comparison with State-of-the-art Methods}

As shown in Table~\ref{tab:main}, our proposed method consistently outperforms existing state-of-the-art methods across all tasks. Specifically, under both $K=8$ and $K=16$ settings, our approach surpasses CONDENSER, a single prompt fusion method. Notably, we achieve a significant relative performance improvement of 5.6\% on the segmentation task. Furthermore, to evaluate the generalization capability of our method, we followed the setup in~\cite{zhang2024instruct,sun2025exploring,wang2025embracing} and constructed a COCO-5$^i$ dataset from MSCOCO~\cite{lin2014microsoft}, ensuring class consistency with Pascal-5$^i$. We then trained our model on COCO-5$^i$ and tested it on Pascal-5$^i$ to verify its cross-dataset performance. The results, presented in Table~\ref{tab:coco}, demonstrate that our method comprehensively outperforms CONDENSER and other existing approaches in this challenging setting. The success of these experiments demonstrates the effectiveness of our method, validating our view that multi-combination collaborative fusion is required to unlock the full potential of multiple contexts.

\subsection{Ablation Study and Qualitative Results}
\subsubsection{Ablation Study for MPGS}
\label{sec:ablation_mpgs}

We conduct a series of ablation studies to validate MPGS, fixing $K=16$ for all ablations. To verify the DRL-inspired disentanglement strategy, we compare different grouping configurations in which the similarity-sorted support set is partitioned into 1, 2, 4, 8, and 16 auxiliary groups with 1 holistic main group. Group=1 fuses all support pairs into a single branch, while the other configurations split them evenly into multiple parallel branches. As shown in Table~\ref{tab:group}, our dual-branch model (Group=2) achieves the highest performance. This confirms that the benefit comes not simply from splitting, but from the principled DRL-inspired separation into high- and low-similarity branches. We further determine the optimal size of each branch. We first vary the size of the high-similarity group $K_{g_1}$ while fixing the low-similarity group parallel to the grouping study above, and observe from Figure~\ref{subfig:g_1} that performance peaks at $K_{g_1}=8$. We then fix $K_{g_1}=8$ and vary $K_{g_2}$, finding a symmetric peak at $K_{g_2}=8$ in Figure~\ref{subfig:g_2}. These results indicate that too few pairs fail to capture sufficient signal, whereas too many dilute the core variation or introduce noise. We therefore set both $K_{g_1}$ and $K_{g_2}$ to 8.

\subsubsection{Ablation Study for MULTI-VQGAN}
\label{sec:ablation_vqgan}
We perform a detailed ablation on MULTI-VQGAN (Table~\ref{tab:ABLATION}). Full model (1) serves as the baseline. The contributions of the individual guidance branches are examined in (2) and (3). Using only the high-similarity branch ($g_1$) or only the low-similarity branch ($g_2$) both lead to reduced performance, indicating that the two branches are complementary and their synergy is essential. In (4) and (5), we further swap the roles of the holistic main branch with either $g_1$ or $g_2$, forcing one of them to act as the foundational representation. Both variants perform worse, confirming that the complete, unfiltered main branch is necessary as the stable base for guidance. Replacing both $g_1$ and $g_2$ with random noise (6) causes a notable performance drop, demonstrating that the improvements come from meaningful learned cues rather than architectural wiring. Finally, freezing the MAE-VQGAN (7) notably degrades performance, indicating the necessity of adapting the backbone to the FUSE module.

Additionally, we conduct ablations for internal architecture of the FUSE module, the core component for integrating multi-faceted signals. Sharing a single MLP for Q/K/V (8) or partially sharing K/V projections (9) reduces performance, confirming that separate MLPs are necessary to learn distinct query matching and value representation roles. Removing cross-attention and averaging guidance (10) further degrades results, demonstrating the importance of dynamic, query-dependent fusion rather than static aggregation. Removing the residual connection (11) also lowers mIoU, indicating that the residual pathway stabilizes learning by framing the output as a refinement rather than a full reconstruction. Overall, separate Q/K/V projections, cross-attention, residual connection work synergistically for effective integration of disentangled guidance signals.

We investigate the optimal layers for injecting guidance features into the MAE-VQGAN encoder. As shown in Figure~\ref{subfig:g_3}, performance peaks around the 11th block, supporting our hypothesis based on \cite{johnson2016perceptual}: mid-layer fusion best balances high-level structural control and low-level detail refinement. Fixing this point, we vary the insertion range and find in Figure~\ref{subfig:g_4} that 7 consecutive layers perform best, showing that guidance should cover a broad region without excessive feature dilution. Therefore, the fusion is applied to a continuous range of intermediate layers, with $N_{\text{down}}=8$ and $N_{\text{up}}=14$.

\subsubsection{Qualitative Results}
Figure~\ref{fig:visual} shows qualitative comparisons on few-shot segmentation, detection, and coloring. Our method better preserves object integrity and fine details, and produces more accurate and tighter detections, especially in cluttered scenes. These results demonstrate that MPGS and MULTI-VQGAN learn more discriminative and robust representations, leading to improved visual predictions.

\section{Conclusions}
\label{sec:formatting}

In this paper, we introduce a novel framework that advances the “retrieve-then-prompt” paradigm in VICL by moving beyond single or fused prompts. Our novel framework uses MPGS strategy to to disentangle prompts into three complementary branches, which are then collaboratively fused by MULTI-VQGAN architecture to generate more robust and accurate predictions.
Extensive experiments on few-shot segmentation, detection, and colorization demonstrate superior cross-task generalization and higher prediction accuracy compared to existing methods. This framework provides a foundation for future work, highlighting the potential of collaborative approaches over traditional prompt selection.

\bibliographystyle{plain}
\newpage
\bibliography{reference}

\clearpage        
\appendix         
\section*{Appendix} 

\input{sup}

\end{document}

%% file: sup.tex
\crefname{section}{Sec.}{Secs.}
\Crefname{section}{Section}{Sections}
\Crefname{table}{Table}{Tables}
\crefname{table}{Tab.}{Tabs.}

\def\cvprPaperID{***} 
\def\confName{CVPR}
\def\confYear{2026}


\section{Overview}

The Supplementary Material is organized as follows:
\begin{itemize}
    \item Section~\ref{sec:rela} provides a comprehensive review of related works.
    \item Section~\ref{sec:MAE} presents preliminary of base inpainting model MAE-VQGAN.
    \item Section~\ref{sec:data} offers detailed information on the datasets used and the specifics of our implementation.
    \item Section~\ref{sec:qu} provides qualitative results to visually demonstrate the performance and versatility of our proposed model.
\end{itemize}

\section{Related Works}
\label{sec:rela}

\subsection{Visual In-Context Learning}
In-Context Learning (ICL), first introduced by GPT-3 \cite{brown2020language}, reformulates NLP tasks as prompt-based text completion. Without parameter updates, LLMs can quickly adapt to new reasoning tasks or unseen patterns by imitating a few in-context prompts, effectively serving as a few-shot learning method \cite{lu2021fantastically,wu2022self,wei2022chain}. Its success has extended to multi-modal domains. For instance, Flamingo \cite{alayrac2022flamingo} expands LLM inputs to include images and videos while keeping language as the interface, enabling tasks like image captioning and visual question answering through mixed text-visual prompts \cite{achiam2023gpt,team2023gemini,team2024gemini}.

Inspired by the success of ICL in language and multi-modal domains, recent research extends this paradigm to the pure vision field with Visual In-Context Learning (VICL), which aims to build a general vision model that performs arbitrary tasks at inference using only a few visual prompts. Pioneering works such as MAE-VQGAN \cite{bar2022visual}, Painter \cite{wang2023images}, and SegGPT \cite{wang2023seggpt} unify diverse visual tasks as inpainting or masked image modeling. Building on this, follow-up studies explore how to improve VICL, especially in visual prompt selection \cite{sun2025exploring,balazevic2023towards}, with methods like VPR \cite{zhang2023makes}, InMeMo \cite{zhang2024instruct}, and Partial2Global \cite{xu2024towards}. Despite these advancements, research on multi-prompt strategies remains relatively underexplored. CONDENSER \cite{wang2025embracing} takes an important step by introducing multi-prompt fusion, its modular design decouples the fusion process from the inpainting stage. However, we contend that a more integrated, collaborative fusion mechanism that jointly leverages diverse contextual cues is necessary to fully unlock the potential of multi-prompt information.

\subsection{Disentangled Representation Learning}

To overcome the limitation that end-to-end deep learning models tend to exploit superficial correlations rather than learning meaningful object properties \cite{geirhos2020shortcut}, Disentangled Representation Learning (DRL) has been proposed \cite{bengio2013representation} and increasingly studied. DRL aims to isolate explanatory latent factors of variation within observed data, yielding semantically meaningful and interpretable representations \cite{bengio2013representation,lake2017building}.

Research on DRL has expanded rapidly, with representative approaches based on generative models \cite{higgins2017beta,chen2016infogan,kingma2013auto,goodfellow2014generative}, as well as causal inference \cite{suter2019robustly} and group theory \cite{higgins2018towards}. These methods encourage latent factor separation while jointly optimizing a primary objective (e.g., generation or discrimination). Due to its capacity to learn explainable, controllable, and robust representations, DRL has been applied in computer vision \cite{lee2018diverse,zhu2018visual,gonzalez2018image,lee2021dranet,liu2021smoothing}, natural language processing \cite{he2017unsupervised,bao2019generating,cheng2020improving}, recommender systems \cite{wang2020disentangled,wang2021multimodal,zhang2020content,ma2019learning}, and graph learning \cite{wang2020disentangled,ma2019disentangled}, improving performance in diverse downstream tasks.

In real-world settings, generative factors often follow hierarchical structures \cite{ross2021benchmarks,li2020progressive}, where factors may be dependent across abstraction levels \cite{ross2021benchmarks} or independent \cite{li2020progressive}. To model this, Hierarchical DRL have been developed, including FineGAN for hierarchical code-based generation \cite{singh2019finegan}, a tree-structured image translation framework \cite{li2021image}, and latent factor activation conditioned on parent values \cite{ross2021benchmarks}. While flat DRL can partially separate multi-level factors, hierarchical designs achieve superior performance, motivating practitioners to consider whether hierarchical dependencies can be exploited for improved disentanglement; therefore, this paper adopts a hierarchical approach.

\begin{figure}
  \centering
  \includegraphics[width=\linewidth]{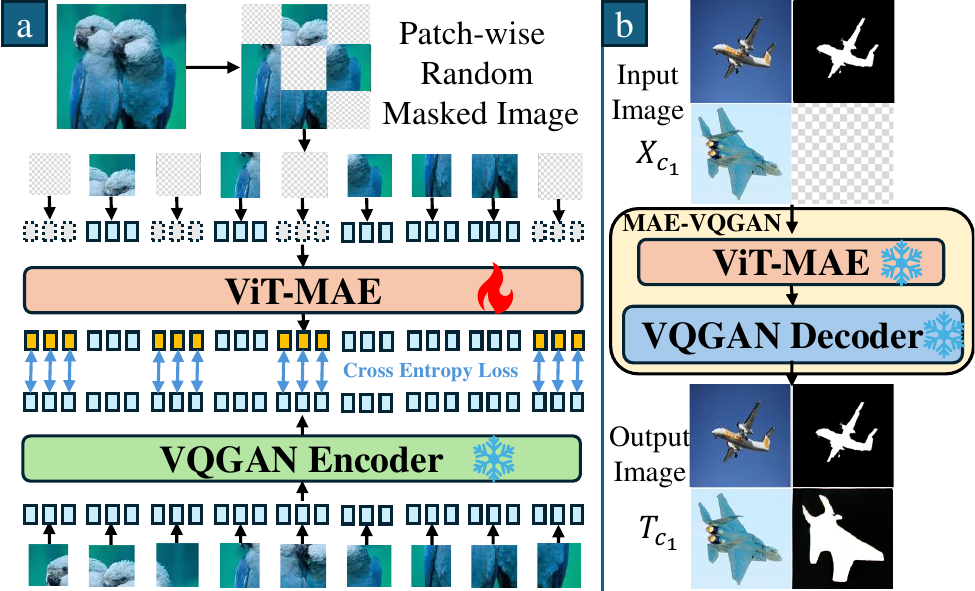} 
  \caption{\textbf{MAE-VQGAN Overview.}
\textbf{(a) Training:} ViT-MAE learns to predict tokens for masked image patches under the supervision of a frozen VQGAN encoder.
\textbf{(b) Inference:} Given a masked input, the trained ViT-MAE predicts the missing tokens, which are then fed into the VQGAN decoder.}
  \label{fig:mae}
\end{figure}

\section{Preliminary of MAE-VQGAN}
\label{sec:MAE}

\label{sec:mae-vqgan}
In this paper, we adopt the image inpainting model MAE-VQGAN proposed by Bar et al.~\cite{bar2022visual}, which has served as the foundation for several existing works~\cite{sun2025exploring, xu2024towards, zhang2024instruct, zhang2023makes}. The model's architecture is composed of an encoder based on the Vision Transformer (ViT)~\cite{vaswani2017attention, dosovitskiy2020image}, which is trained using a Masked Autoencoder (MAE)~\cite{he2022masked} objective (denoted as ViT-MAE), and a decoder taken from a Vector Quantised Generative Adversarial Network (VQGAN)~\cite{esser2021taming}.

As illustrated in Figure~\ref{fig:mae}(a), the training process follows the MAE paradigm. The ViT-MAE encoder learns to reconstruct an image from its unmasked patches by predicting the corresponding discrete tokens in the VQGAN's codebook. The predicted tokens are then optimized using a cross-entropy loss against the ground-truth tokens generated by the VQGAN's encoder.

During inference, depicted in Figure~\ref{fig:mae}(b), the trained ViT-MAE takes the constructed input image \(X_{c_1}\) and generates a complete set of predicted tokens. These tokens are subsequently passed to the VQGAN decoder, which synthesizes them into the final output image \(T_{c_1}\). The overall operation of the MAE-VQGAN model, denoted as \(\varphi\), can be summarized as:
\begin{equation}
  \label{eq:mae-vqgan}
  T_{c_1} = \varphi(X_{c_1}).  
\end{equation}

\section{Datasets and Implementation Details}
\label{sec:data}

\subsection{Performance Evaluation}
To evaluate model performance on different downstream tasks, we use the following standard metrics:
\begin{itemize}
\item \textbf{Foreground segmentation} and \textbf{single object detection}: measured using \textbf{mean Intersection-over-Union (mIoU)}, which compares the overlap between predicted and ground-truth masks or bounding boxes.
\item \textbf{Image colorization}: evaluated using \textbf{Mean Squared Error (MSE)} between the predicted and ground-truth pixel values.
\end{itemize}

\subsection{Datasets and Tasks}

Our experiments are conducted using three widely adopted public datasets: PASCAL-5$^i$, PASCAL VOC 2012, and ImageNet-1K, which we adapt for three downstream tasks: foreground segmentation, single-object detection, and image colorization.

PASCAL-5$^i$~\cite{shaban2017one} is a few-shot segmentation benchmark derived from PASCAL VOC 2012, where 20 categories are partitioned into four disjoint folds. Following the sample configuration from~\cite{zhang2024instruct}, we use 2,286, 3,425, 5,883, and 2,086 samples for the four folds.

We further utilize PASCAL VOC 2012~\cite{everingham2015pascal} for evaluating single-object detection. To ensure experimental clarity, we retain only images that contain a single annotated object and train using 612 in-context samples.

Finally, ImageNet-1K~\cite{russakovsky2015imagenet} is used to explore image colorization. We randomly sample 50K images from the 1.2M training set, convert them to grayscale, and train the model to reconstruct the corresponding original color images. Quantitative evaluation is performed on the official validation set using MSE to measure pixel-wise reconstruction error.

\subsection{Implementation Details}
For a fair comparison, the pretraining protocol for MAE-VQGAN follows \cite{bar2022visual} and \cite{sun2025exploring}, while the training procedure for the Prompt Generator is aligned with \cite{wang2025embracing}.

In the training of our MULTI-VQGAN, we set the learning rate to 0.05 and trained the model for 10 epochs with a batch size of 16. To validate the benefits of our approach across different configurations, we selected \(K\) values of 8 and 16. For the MPGS process, the corresponding \(\{K_{g_1}, K_{g_2}\}\) pairs were set to \{4, 4\} for \(K=8\) and \{8, 8\} for \(K=16\). In the Hierarchical Fusion stage, the parameters \(N_{\text{down}}\) and \(N_{\text{up}}\) were set to 8 and 14, respectively. The entire training and experimentation process was conducted on a single V100 GPU. More experimental details are provided in the supplementary material.

\section{Qualitative Results}
\label{sec:qu}

\begin{figure*}
    \centering

    \textbf{Qualitative Results for Segmentation}\par\medskip
    \begin{subfigure}{0.19\textwidth}\includegraphics[width=\linewidth]{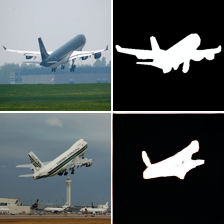}\end{subfigure}\hfill
    \begin{subfigure}{0.19\textwidth}\includegraphics[width=\linewidth]{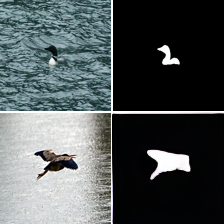}\end{subfigure}\hfill
    \begin{subfigure}{0.19\textwidth}\includegraphics[width=\linewidth]{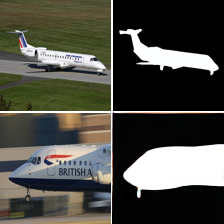}\end{subfigure}\hfill
    \begin{subfigure}{0.19\textwidth}\includegraphics[width=\linewidth]{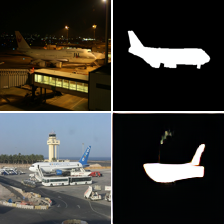}\end{subfigure}\hfill
    \begin{subfigure}{0.19\textwidth}\includegraphics[width=\linewidth]{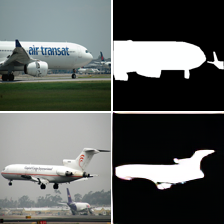}\end{subfigure}
    \vspace{2pt} \\
    \begin{subfigure}{0.19\textwidth}\includegraphics[width=\linewidth]{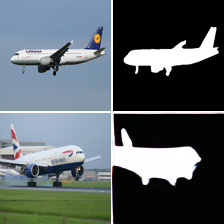}\end{subfigure}\hfill
    \begin{subfigure}{0.19\textwidth}\includegraphics[width=\linewidth]{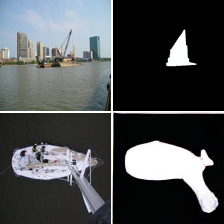}\end{subfigure}\hfill
    \begin{subfigure}{0.19\textwidth}\includegraphics[width=\linewidth]{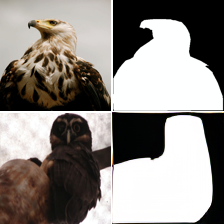}\end{subfigure}\hfill
    \begin{subfigure}{0.19\textwidth}\includegraphics[width=\linewidth]{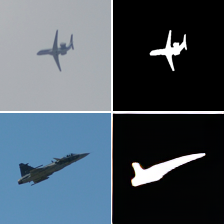}\end{subfigure}\hfill
    \begin{subfigure}{0.19\textwidth}\includegraphics[width=\linewidth]{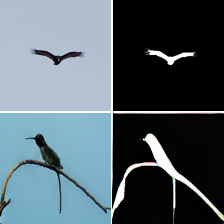}\end{subfigure}
    

    \textbf{Qualitative Results for Detection}\par\medskip
    \begin{subfigure}{0.19\textwidth}\includegraphics[width=\linewidth]{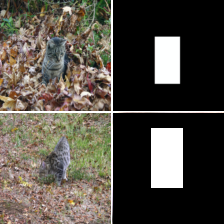}\end{subfigure}\hfill
    \begin{subfigure}{0.19\textwidth}\includegraphics[width=\linewidth]{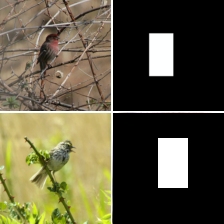}\end{subfigure}\hfill
    \begin{subfigure}{0.19\textwidth}\includegraphics[width=\linewidth]{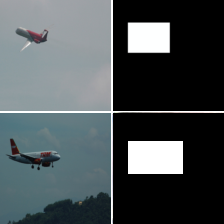}\end{subfigure}\hfill
    \begin{subfigure}{0.19\textwidth}\includegraphics[width=\linewidth]{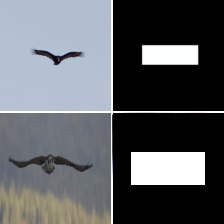}\end{subfigure}\hfill
    \begin{subfigure}{0.19\textwidth}\includegraphics[width=\linewidth]{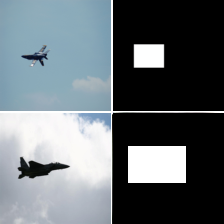}\end{subfigure}
    \vspace{2pt} \\
    \begin{subfigure}{0.19\textwidth}\includegraphics[width=\linewidth]{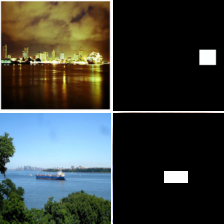}\end{subfigure}\hfill
    \begin{subfigure}{0.19\textwidth}\includegraphics[width=\linewidth]{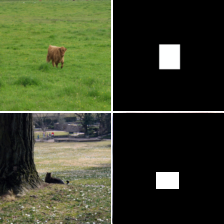}\end{subfigure}\hfill
    \begin{subfigure}{0.19\textwidth}\includegraphics[width=\linewidth]{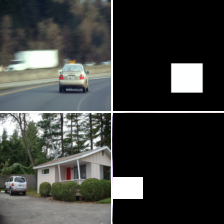}\end{subfigure}\hfill
    \begin{subfigure}{0.19\textwidth}\includegraphics[width=\linewidth]{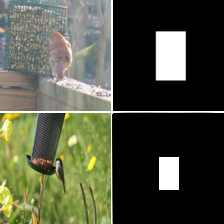}\end{subfigure}\hfill
    \begin{subfigure}{0.19\textwidth}\includegraphics[width=\linewidth]{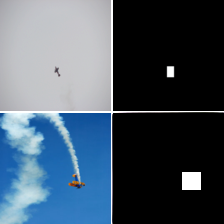}\end{subfigure}
    \textbf{Qualitative Results for Coloring}\par\medskip
    \begin{subfigure}{0.19\textwidth}\includegraphics[width=\linewidth]{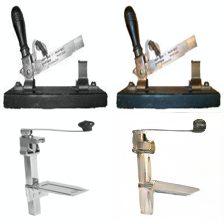}\end{subfigure}\hfill
    \begin{subfigure}{0.19\textwidth}\includegraphics[width=\linewidth]{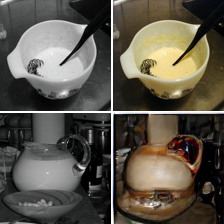}\end{subfigure}\hfill
    \begin{subfigure}{0.19\textwidth}\includegraphics[width=\linewidth]{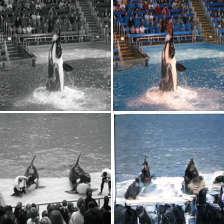}\end{subfigure}\hfill
    \begin{subfigure}{0.19\textwidth}\includegraphics[width=\linewidth]{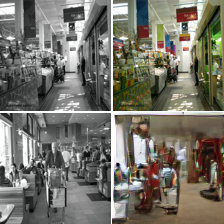}\end{subfigure}\hfill
    \begin{subfigure}{0.19\textwidth}\includegraphics[width=\linewidth]{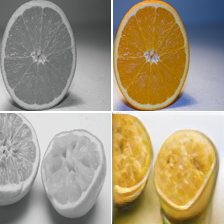}\end{subfigure}
    \vspace{2pt} \\
    \begin{subfigure}{0.19\textwidth}\includegraphics[width=\linewidth]{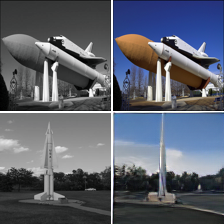}\end{subfigure}\hfill
    \begin{subfigure}{0.19\textwidth}\includegraphics[width=\linewidth]{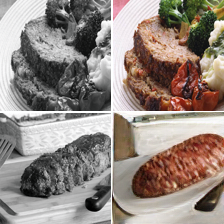}\end{subfigure}\hfill
    \begin{subfigure}{0.19\textwidth}\includegraphics[width=\linewidth]{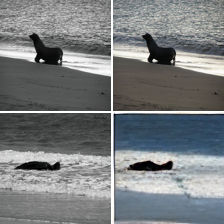}\end{subfigure}\hfill
    \begin{subfigure}{0.19\textwidth}\includegraphics[width=\linewidth]{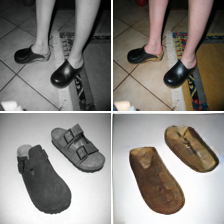}\end{subfigure}\hfill
    \begin{subfigure}{0.19\textwidth}\includegraphics[width=\linewidth]{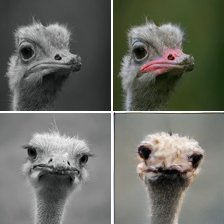}\end{subfigure}

    \caption{Qualitative results for our model on three different downstream tasks. From top to bottom: Segmentation, Detection, and Coloring, showcasing the model's ability to generate high-quality, coherent outputs across diverse applications.}
    \label{fig:qualitative_all_tasks}
\end{figure*}

In this section, we present a series of qualitative results (as shown in Figure~\ref{fig:qualitative_all_tasks}) to visually demonstrate the powerful performance and versatility of our model, serving as a compelling supplement to the quantitative metrics.

